\def\year{2020}\relax
\documentclass[letterpaper]{article} 
\usepackage{aaai20}  
\usepackage{times}  
\usepackage{helvet} 
\usepackage{courier}  
\usepackage[hyphens]{url}  
\usepackage{graphicx} 
\usepackage{graphicx}  
\usepackage{amsmath}
\title{Self-supervised Scale Equivariant Network for\\ Weakly Supervised Semantic Segmentation}
\author{Yude Wang\textsuperscript{\rm 1,2}, Jie Zhang\textsuperscript{\rm 1}, Meina Kan\textsuperscript{\rm 1}, Shiguang Shan\textsuperscript{\rm 1,2,3}, Xilin Chen\textsuperscript{\rm 1,2}\\
	\textsuperscript{\rm 1}Key Lab of Intelligent Information Processing of Chinese Academy of Sciences (CAS),\\Institute of Computing Technology, CAS, Beijing, 100190, China\\
	\textsuperscript{\rm 2}University of Chinese Academy of Sciences, Beijing, 100049, China\\
	\textsuperscript{\rm 3}CAS Center for Excellence in Brain Science and Intelligence Technology, Shanghai 200031, China\\
	\{yude.wang, jie.zhang\}@vipl.ict.ac.cn, \{kanmeina, sgshan, xlchen\}@ict.ac.cn
}

\begin{document}
	
	\maketitle
	
	\begin{abstract}
	Weakly supervised semantic segmentation has attracted much research interest in recent years considering its advantage of low labeling cost. Most of the advanced algorithms follow the design principle that expands and constrains the seed regions from class activation maps (CAM). As well-known, conventional CAM tends to be incomplete or over-activated due to weak supervision. Fortunately, we find that semantic segmentation has a characteristic of spatial transformation equivariance, which can form a few self-supervisions to help weakly supervised learning. This work mainly explores the advantages of scale equivariant constrains for CAM generation, formulated as a self-supervised scale equivariant network (SSENet). Specifically, a novel scale equivariant regularization is elaborately designed to ensure consistency of CAMs from the same input image with different resolutions. This novel scale equivariant regularization can guide the whole network to learn more accurate class activation. This regularized CAM can be embedded in most recent advanced weakly supervised semantic segmentation framework. Extensive experiments on PASCAL VOC 2012 datasets demonstrate that our method achieves the state-of-the-art performance both quantitatively and qualitatively for weakly supervised semantic segmentation. Code has been made available\footnote{https://github.com/YudeWang/SSENet-pytorch}.
	\end{abstract}
	
	\section{Introduction}
	Deep convolutional neural networks have achieved remarkable successes in recent years with the support of massive labeled data. While the research moves forward slowly burdened with expensive data annotation processes, especially for the semantic segmentation, whose annotation requirement is much more complex and expensive than the classification and detection tasks. Therefore, some weakly supervised semantic segmentation works focus on training network with lower-level supervision, such as bounding boxes~\cite{dai2015boxsup,khoreva2017simple}, scribbles~\cite{lin2016scribblesup,vernaza2017learning} and points~\cite{bearman2016s}, which are much cheaper to be labeled than pixel-level segmentation mask. Image-level category label is the most common used supervision since there are already many large-scale classification datasets, such as ImageNet~\cite{deng2009imagenet}. To the best of our knowledge, almost all the research of image-level weakly supervised semantic segmentation methods are based on Class Activation Maps (CAM)~\cite{zhou2016learning}, which generate a roughly activated feature map to locate objects spatial positions. The original version of CAM has relatively complete coverage on small objects, while with the increasing of object size, the activated regions shrink into the most discriminative part, e.g. the head of a dog and the wheel of a car. The incomplete activation maps as pseudo segmentation labels heavily damage the training of segmentation network, leading to severe performance degradation.
	\begin{figure}[t]
		\centering
		\includegraphics[width=1.0\columnwidth]{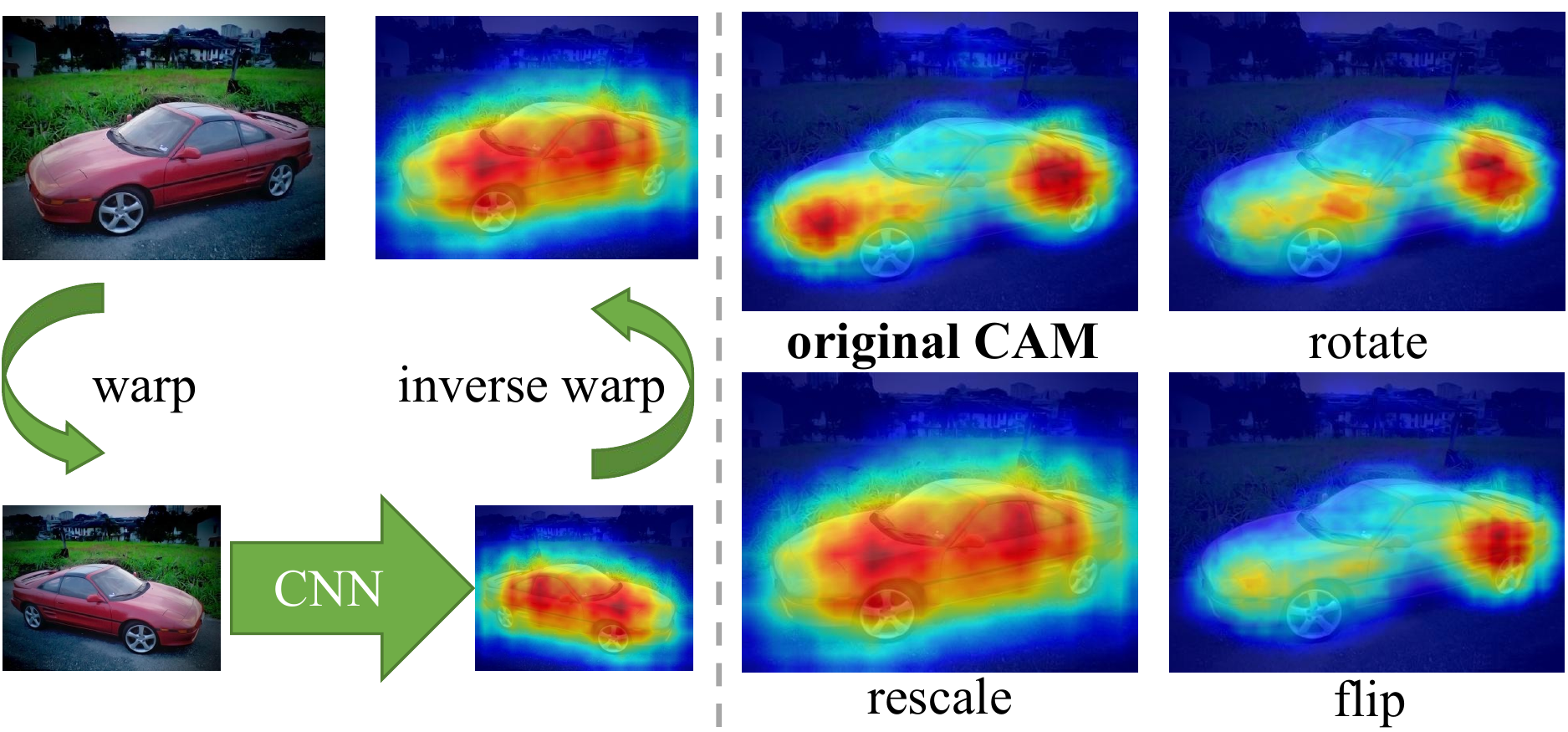}
		\caption{Comparing the inconsistency between original CAM with other CAMs warped by various spatial transformations, following the pipeline on the left.}
		\label{fig:affine}
	\end{figure}
	
    The ideal segmentation network should be an affine transformation equivariant mapping function. As shown in Fig.~\ref{fig:affine}, the input image is warped by some affine transformations and then fed into CNN to achieve CAMs. The generated CAMs are then inverse warped to meet with the CAM generated from the original image. The wrapped CAMs should keep the same with the original CAM. However, Fig.~\ref{fig:affine} illustrates that there is an inconsistency between the wrapped CAMs and the origin one, especially by the rescaling operation. The equivariant constraint has been implicitly used in fully supervised segmentation that the pixel-level labels always keep consistent with corresponding pixels during spatial transformation augmentation. However, the class activation maps learned by image-level supervision are not equivariant in most cases, which may hurt the generalization performance of CAM based weakly supervised semantic segmentation methods.

	Considering that the conventional CAM has severe inconsistency by rescaling transformation that it covers more background regions on small object and covers fewer foreground regions on large objects, we turn to regularize the class activation map of different scales to refine each other. In this paper, we propose a novel two branch self-supervised scale equivariant network (SSENet) to overcome the drawback of CAM mentioned above by a self-supervision framework. The network constrains the activated feature maps to be scale equivariant, i.e. the activation of images keep consistent on various scales. With the scale equivariant regularization (SER), the CAM consistency is significantly improved as shown in Fig.~\ref{fig:cams}. The regularization is effective and easy to be employed on any CAM-based algorithm for weakly supervised semantic segmentation. Benefited from the improved CAM, the performance of weakly supervised semantic segmentation will be further improved. 

	The main contributions can be summarized as follows:
	\begin{itemize}
	\item We propose a novel scale equivariant regularization (SER) to narrow the consistency gap between the CAMs generated from various scale images, leading to significant improvement on CAMs.
	\item We propose a novel self-supervised scale equivariant network (SSENet) architecture, which is the first try to utilize self-supervised learning for image-level weakly supervised semantic segmentation.
	\item Experiments on PASCAL VOC 2012 dataset~\cite{everingham2015pascal} demonstrate the outstanding performance of our SSENet comparing with state-of-the-arts.
	\end{itemize}
	
	\section{Related Work}

	\subsection{Weakly Supervised Semantic Segmentation}

	Although fully supervised semantic segmentation algorithms~\cite{chen2014semantic,Long_2015_CVPR} have achieved great successes in recent years, the pixel-level annotations are expensive to collect, resulting in that more and more weakly supervised approaches are proposed and studied to alleviate this practical problem.
	
	Image-level supervision means only object category labels are available during training time. The most fundamental work CAM~\cite{zhou2016learning}, trains an image classification network and multiples the weight of fully connect layer on the feature map during inference for roughly object localization. Early approaches adopt various strategies, such as EM algorithm~\cite{papandreou2015weakly} and multiple instance learning~\cite{pinheiro2015image} to achieve pseudo segmentation labels. SEC~\cite{kolesnikov2016seed} proposes three principles for the task, which selects confident initial seeds from CAM, expands activate regions by global weighted rank pooling and constrains segmentation boundary considering color information. The CAM generation network always activates on the discriminative parts of the objects, remaining a challenge to predict segmentation mask covering the entire foreground object. The adversarial erasing strategy is widely employed in~\cite{wei2017object,hou2018self} to solve this case by erasing discriminative regions and mining others. FickleNet~\cite{lee2019ficklenet} randomly dropout the weight of convolution at each position to discover more class activated regions. Another expanding method from reliable seed regions is random walk with transition matrix~\cite{ahn2018learning,ahn2019weakly}. The matrix can be derived from AffinityNet supervised by classified pixel pair~\cite{ahn2018learning} or learned from the class boundary map~\cite{ahn2019weakly}.
	
	\subsection{Self-supervised learning}
	
	Comparing to fully supervised network training with massive annotated data, self-supervised learning is a candidate solution to learn more robust visual features without any additional annotation cost. Most of the self-supervised learning algorithms propose pretext tasks, which are predefined to generate controlled supervision signals from the input domain as optimization directions for deep neural networks. The task-related feature representations will be learned through this process~\cite{jing2019self}. To some extent, These self-supervised pre-trained features bring comparable performance improvement with ImageNet pre-trained model~\cite{doersch2015unsupervised}.
	
	Here are some self-supervised research works based on various pretext tasks. The most famous ones are generative models, e.g. generative adversarial networks~\cite{goodfellow2014generative}. These models learn the feature distribution from a large unlabeled dataset by adversarial training. There are also some discriminative models learning with pretext tasks on static images, e.g. relative position prediction~\cite{doersch2015unsupervised} and spatial transformation prediction~\cite{gidaris2018unsupervised}.
	
	\subsection{Scale Equivariance}
	
	As for the research of scale invariance and scale equivariance, most of the works focus on designing special network architecture to preserve the scale invariance or scale equivariance~\cite{Kanazawa2014locally,worrall2019deep}. Since the sizes of convolution kernels are discrete, it is hard to perfectly achieve scale equivariance by refining network architecture. It still leaves a long way to go. In this paper, our proposed SSENet resorts to constraining the activation map consistency on various scales during network training instead of designing the equivariant module. The proposed framework effectively preserves the scale equivariance of CAM, which significantly improves the generated pseudo labels for weakly supervised semantic segmentation problem.
	\begin{figure*}[t]
		\centering
		\includegraphics[width=0.85\textwidth]{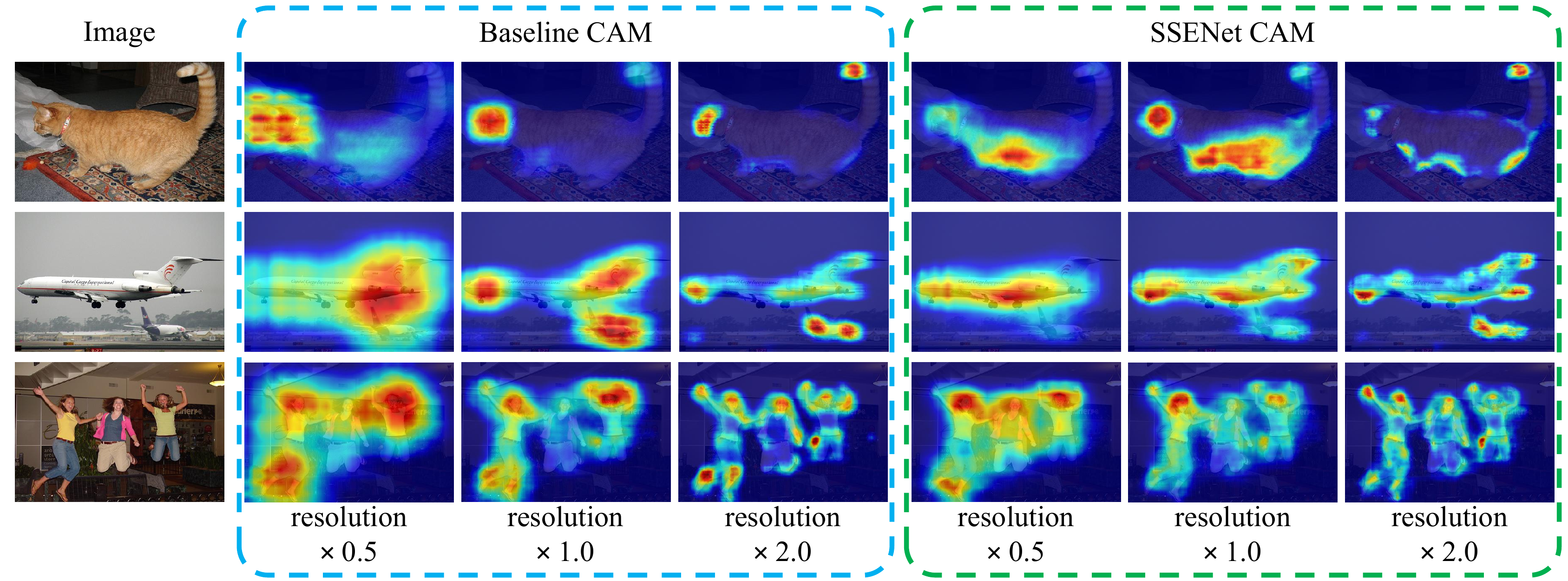}
		\caption{Visualization of CAMs on various scales. The left column shows the original images. The middle part shows the conventional CAMs and the right part is the CAMs generated by our SSENet. As shown in the figure, the CAMs generated by our SSENet have better consistency on various scales.}
		\label{fig:cams}
	\end{figure*}
	
	\section{Approach}
	
	In this section, we will carefully present the idea of the proposed two-branch self-supervised scale equivariant network (SSENet). Firstly we will illustrate the scale inconsistency problem of class activation map in weakly supervised semantic segmentation and analyze the essential cause of it. Secondly, we will describe the details of the self-supervised regularization which improves the scale consistency for class activation map. Finally, we will show how to integrate scale equivariant regularization into a two-branch network for weakly supervised semantic segmentation.
	
	\subsection{Observation}
	
	To figure out the semantic regions in a static image, most of the approaches follow the work~\cite{zhou2016learning} by training a CNN with the image-level label to get class activation map (CAM). It is well known to all that CAMs always located in the most discriminative part, especially for the large objects. It is hard to cover the entire object regions. Meanwhile, as shown in Fig.~\ref{fig:cams}, CAM highlights a relatively expanded activation map for small object covering too many background regions. The inconsistency extremely damages the network performance for solving weakly supervised semantic segmentation problem. Although researches are focusing on finding more discriminative regions, e.g. adversarial erasing strategies~\cite{wei2017object} remove the most highlighted parts to activate more class-related feature regions. While it remains a challenge to reduce those over-activated background regions around small objects. Comparing with fully supervised semantic segmentation, image-level supervision is too weak to determine the object boundaries, additional supervisions or regularizations should be employed on the network to elevate performance.
	
	For the reason that the training process of CAM always excludes the background category, it becomes a tricky problem to generate pseudo segmentation mask which contains the background. One of the simplest methods is using a hard threshold to separate the foreground and the background. However, as shown in Fig.~\ref{fig:cams} that the best threshold of CAM is different on various scales. The threshold parameter should be large enough to exclude those over-activated regions on small objects, while it should be relatively small to include more activation parts on incomplete CAM for large objects. Although there are some post-processing approaches such as dense CRF~\cite{chen2018deeplab} fine-tuning the contour of activation regions to some extent by color constraints, it can not solve the problem fundamentally. It is hoped that the contour should be well determined by a scale equivariant activation map, making it more general for objects of various scales. Therefore, it is reasonable to employ self-supervised scale equivariance constrain as auxiliary supervision to help image-level weakly supervised network training.
	\begin{figure*}[htbp]
		\centering
		\includegraphics[width=0.8\textwidth]{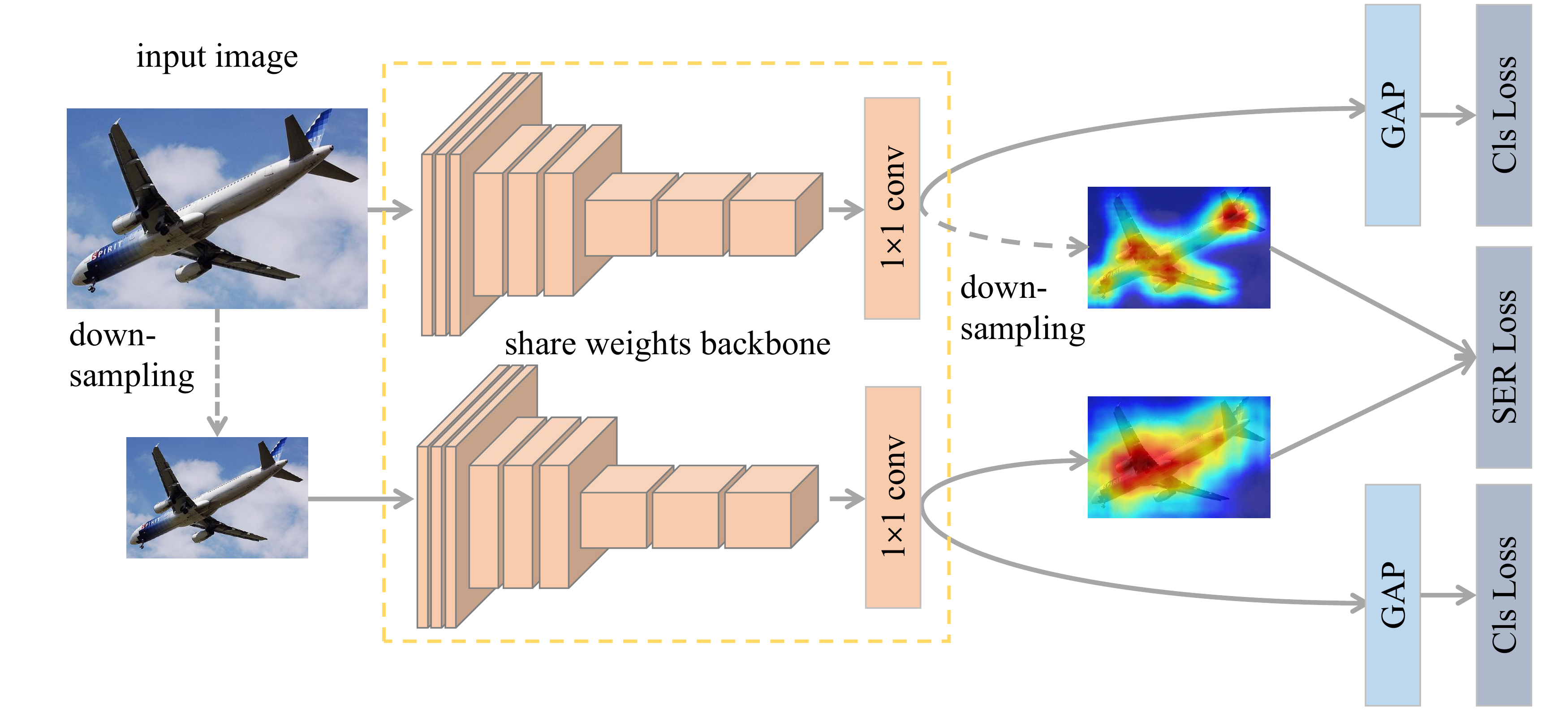}
		\caption{The two-branch architecture of SSENet. The SSENet takes the original and the downsampled version of images as input, endeavoring to preserve the consistency of CAMs from each branch by scale equivariant regularization loss (\textbf{SER Loss}). \textbf{GAP} denotes the global average pooling layer. \textbf{Cls Loss} denotes the multi-label classification loss.}
		\label{fig:network}
	\end{figure*}
	
	\subsection{Self-supervised Regularization}
	
	The fully supervised semantic segmentation task can be generally formulated as $f_{\theta_s}(x)=y$, where $x$ denotes the input image with the corresponding segmentation ground truth mask $y$. $f_\theta(\cdot)$ denotes the CNN based nonlinear mapping function. While for the image-level weakly supervised semantic segmentation, the network is attached with additional global pooling function $P(\cdot)$ to solve classification task as $P(f_{\theta_c}(x))=l$, where $l$ is image category label of $x$. Ideally, the network parameters $\theta_s$ and $\theta_c$ should keep the same if the fully supervision task and weakly supervision task have the same optimization objective.
	
	Random scale input images is a common practice of data augmentation for network training, which can always improve the inference performance. Suppose there is an affine transformation matrix $A$, the fully supervised segmentation task requires the equivariance of affine transformation as $f_{\theta_s}(Ax)=Ay$, while the weakly supervision task focusing on the mapping invariance as $P(f_{\theta_c}(Ax))=l$. The invariance mainly caused by pooling function $P(\cdot)$, but there is no explicit equivariance constrain for $f_{\theta_c}(\cdot)$. The different optimization objectives make it hard to guarantee the networks with different supervisions have the same convergence point, and it is nearly impossible to achieve $\theta_s$ with only image-level label $l$. Image-level weakly supervised semantic segmentation needs additional supervisions to guide the optimization direction.
	
	Under the scenario of weakly supervised learning, only self-supervised labels and handcrafted constraints are available to narrow the gap of optimal solution between classification and semantic segmentation tasks. We resort to the self-supervised learning, proposing scale equivariant regularization (SER) to improve the CAM quality.
	\begin{equation}\label{eq:lossfunc}
	\min_{\theta}\sum_i \frac{||P(f_{\theta}(x_i))-l_i||+||P(f_{\theta}(Ax_i))-l_i||}{2}+\eta R_i,	
	\end{equation}
	\begin{equation}\label{eq:regularization}
	R_i=||f_{\theta}(Ax_i)-Af_{\theta}(x_i)||.
	\end{equation}
	Our proposed cost function with scale equivariant regularization is given as Eq.~\ref{eq:lossfunc}. $i$ denotes the sample index and $\eta$ is the parameter to control the influence of regularization which is set to $1$ in the following experiments without careful tuning. The formulation of the scale equivariant regularization is defined in Eq.~\ref{eq:regularization}, where $A$ is the matrix form of bilinear interpolation. The elements in matrix $A$ is predefined, e.g. $1/2$ downsampling operation, and fixed in the training process. Since the input images are randomly rescaled during data augmentation preprocessing, SER endeavors to preserve the scale equivariance on the overall scale range.
	
    Moreover, the extended version of scale equivariant regularization is that $A$ can be any spatial linear transformation, e.g. flip and rotation. In order to verify the effectiveness of the proposed equivariant regularization, our work only focuses on scaling inconsistency. As shown in Fig.~\ref{fig:cams}, training with scale equivariant regularization obviously increase the similarity of class activation map on various scales. And the trend of CAM contour turns to meet the object boundary closely. At the same time, the regularized activation maps have more complete coverage of the entire object comparing to the unconstrained ones, which demonstrates that the scale equivariance regularization has the same effect as the methods of object region mining.
	
	\subsection{Self-supervised Scale Equivariant Network}
	
	We propose a novel weight-shared two-branch network to employ the scale equivariant regularization (SER) for network training. As shown in Fig.~\ref{fig:network}, the randomly selected RGB image, and its downsampled copy are fed into the two branch network respectively. At the end of the backbone networks, the $C-1$ channels feature maps are achieved, where $C-1$ is the number of categories excluding the background. And the feature maps are also known as class activation maps. Global average pooling layer is attached to the feature map with multi-label classification loss as supervision. Considering that the scales of CAM from two branch are not the same, we downsample the CAM output from the large branch by bilinear interpolation to keep scale consistent with the one from the other branch. $L_2$ distance is used to measure the gap between these two branch CAMs, working as the scale equivariant regularization to constrain the activation consistency during the training process. Finally, the loss for network training is the weighted sum of classification losses mean and SER item with $\eta=1$ in Eq.~\ref{eq:lossfunc}.
	
	During the test phase, we preserve one branch as the inference network to obtain the final class activation maps. Noting that the two branches of the network have shared weights, it is equivalent to preserve the parameters from the large or small branch. Moreover, the additional background score maps will be concatenated with $C-1$ channel feature maps manually to form preliminary segmentation results. The background score maps are defined as follows.
	\begin{equation}
	M_c=\left\{
	\begin{array}{rcl}
	& \alpha & \text{$c$ is background}\\
	& \frac{ReLU(\hat{y}_c-\epsilon)}{\max ReLU(\hat{y}_c)+\epsilon} & \text{others}
	\end{array} \right.
	\end{equation}
	Where $\hat{y}_c$ denotes the predicted activation map of class $c$ and $\epsilon$ is set as $10^{-5}$ to avoid dividing zero. $M_c$ is the normalized activation map of class $c$. $\alpha$ denotes the parameter used to control the background confidence score which is set to $0.2$ in our experiments. We choose ResNet-38 as the backbone network in all experiments, following the setting of~\cite{ahn2018learning}. Dense CRF~\cite{chen2018deeplab} is attached as a post-processing step to refine the contour of CAM to be more close with the object boundary.
	
    To further improve the performance of weakly supervised semantic segmentation, we follow the pipeline of ~\cite{ahn2018learning}. In shortly, several pixel pairs are sampled based on the improved CAM $(h\times w)$ to train an AffinityNet, and the pixel affinity matrix $(hw\times hw)$ is calculated by the feature map of AffinityNet. Then the improved CAM is resized into $1\times hw$, multiplied with pixel affinity matrix several times, which is named as random walk step, and CAM vector is resized back to $h\times w$ eventually as the pseudo label. Finally, a classical semantic segmentation model DeepLab is trained by these pseudo labels. We will carefully investigate the performance improvement of each step in next section to show the benefits of our SSENet to weakly supervised semantic segmentation.
	
	\section{Experiments}
	
	In order to illustrate the superiority of our proposed SSENet, we conduct some weakly supervised semantic segmentation experiments from several aspects. SSENet is also embedded into advanced weakly supervised semantic segmentation framework without using any additional data, achieving outstanding performance with other state-of-the-art methods. 
	
	\subsection{Implementation Details}
	
	To evaluate the effectiveness of our SSENet, we adopt PASCAL VOC 2012 benchmark~\cite{everingham2015pascal} which contains 20 object categories and another background category. With the additional annotation of SBD~\cite{BharathICCV2011}, the common setting of fully supervised semantic segmentation task takes 10,582 images as the augmented training set, 1,449 for validation and 1,456 for testing. Our experiments follow the same dataset partition while only image-level classification labels are provided during network training, and take the mean intersection of union (mIoU) as evaluation metric as well as other previous works.
	
    As for training settings, the backbone network used in our experiments is the modified version of ResNet38\footnote{Model A1 version in~\cite{wu2019wider}} pretrained on ImageNet, which takes two $3\times3$ convolution layers in each residual block instead of the bottleneck structure, removing original global average pooling and fully connected layers. As well known that ResNet groups several residual blocks into one level, using stride 2 convolution layer at the beginning. We replace the stride convolution of the last two levels by dilation convolution with rate 2 and 4 in last two levels respectively to keep the same network receptive field. Additional $1\times1$ convolution is attached to the end of the network as the pixel-wise classifier, followed by a global average pooling layer which pools the feature map into feature vector for the classification task. The network is trained on 4 Titan-xp GPUs with batch size 8 for 15 epochs. The initial learning rate is 0.01 and the training schedule follows the poly policy that $lr=lr_{init}*(1-\frac{itr}{max\_itr})^{\gamma}$, where $\gamma=0.9$ in our experiments. The input images are randomly rescaled into [448, 768] on the longest edge, then randomly cropped by $448\times448$ and fed into the large branch of the network. The regularization weight in Eq.~\ref{eq:lossfunc} is set as $\eta=1$ in all our experiments. During inference, only one branch network remains since these two branches have shared weights. Flip and multi-scale test are adopted to improve performance. 
	\begin{table}[t]
		\centering
		\begin{tabular}{lcc}
			\hline
			Model & CAM (mIoU) & CAM+rw (mIoU)\\
			\hline
			Baseline & 47.3\% & 58.8\%\\
			\hline
			SSENet (0.6) & 48.5\% & 61.5\% \\
			SSENet (0.5) & 48.9\% & 61.7\%\\
			SSENet (0.4) & 49.4\% & 61.8\%\\
			SSENet (0.3) & \textbf{49.8\%} & \textbf{62.1\%}\\
			SSENet (0.2) & 49.4\% & 61.7\%\\
			\hline
		\end{tabular}
		\caption{Comparison between baseline model and SSENet with various branch downsampling rates (given in parentheses). We evaluate the generated pseudo labels from CAM generation step (\textbf{CAM}) and the following random walk step (\textbf{CAM+rw}) on PASCAL VOC 2012 train set.}
		\label{tab:branch}
	\end{table}
	\begin{table}[t]
		\centering
		\begin{tabular}{lcc}
			\hline
			Model & scaling range &CAM (mIoU)\\
			\hline
			Baseline & [448, 768] & 47.3\%\\
			Baseline & [224, 768] & 46.6\%\\
			SSENet (0.5) & [448, 768] & 48.9\%\\
			\hline
		\end{tabular}
		\caption{Pseudo label comparison between baseline model and SSENet (0.5 downsampling rate) with different scale augmentation ranges on PASCAL VOC 2012 train set.}
		\label{tab:scaleaug}
	\end{table}
	\subsection{Ablation Study}
	\paragraph{Branch Downsampling Rate} Branch downsampling rate of the SSENet is a significant super parameter to control the effectiveness of scale equivariance regularization. We make some experiments on various branch downsampling rate which are evaluated by multi-scale and flip test. Tab.~\ref{tab:branch} shows networks trained with SER work better on the performance of CAM than baseline models over all branch downsampling rates. And the mIoU improvements are further boosted by employing random walk step. When selecting branch downsampling rate as 0.3, the mIoU of CAM produced by SSENet achieves 2.5\% improvement and 3.3\% after random walk step. Besides, they are further utilized as the pseudo labels to train advanced fully-supervised semantic segmentation network, leading to the performance improvement of the whole solution framework.
	\begin{figure*}[t]
		\centering
		\includegraphics[width=1.0\linewidth]{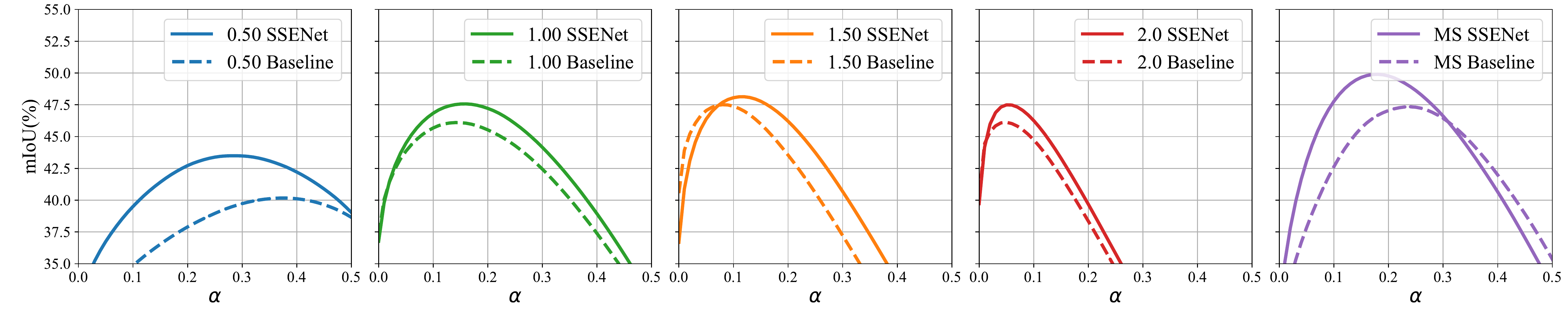}
		\caption{The evaluation of generated pseudo labels on PASCAL VOC 2012 train set by baseline and SSENet. The decimals in the legends are the scaling rate of single-scale test, \textbf{MS} denotes multi-scale test, and $\alpha$ is the background confidence threshold.}
		\label{fig:exp_b}
	\end{figure*}	
	\paragraph{Scale Augmentation Range} Rescaling is a basic data augmentation method for network training, which elevates the robustness of the network to different image scales. In our SSENet, the downsampling branch resizes the input images into a smaller scale, enlarging the scale augmentation range to some extent. To verify whether the performance improvement comes from the larger scaling range, we train the baseline model and SSENet with different scale augmentation range, and the Tab.~\ref{tab:scaleaug} summarizes the experiment results. It shows that SSENet with 0.5 downsampling rate works better than baseline model when scaling range is [448, 768], noting that the scale range of SSENet downsampling branch is [224, 384] at the same time. Besides, we train a baseline model with [224, 768] scale augmentation, which expands the range to include the image scales from SSENet downsampling branch. As seen in the Tab.~\ref{tab:scaleaug}, the baseline model degenerates by 0.7\% mIoU comparing to that trained with [448, 768] scale augmentation, illustrating that the performance contribution of SSENet mainly comes from scale equivariant regularization rather than larger scaling range. 
	\begin{figure}[t]
		\centering
		\includegraphics[width=0.9\columnwidth]{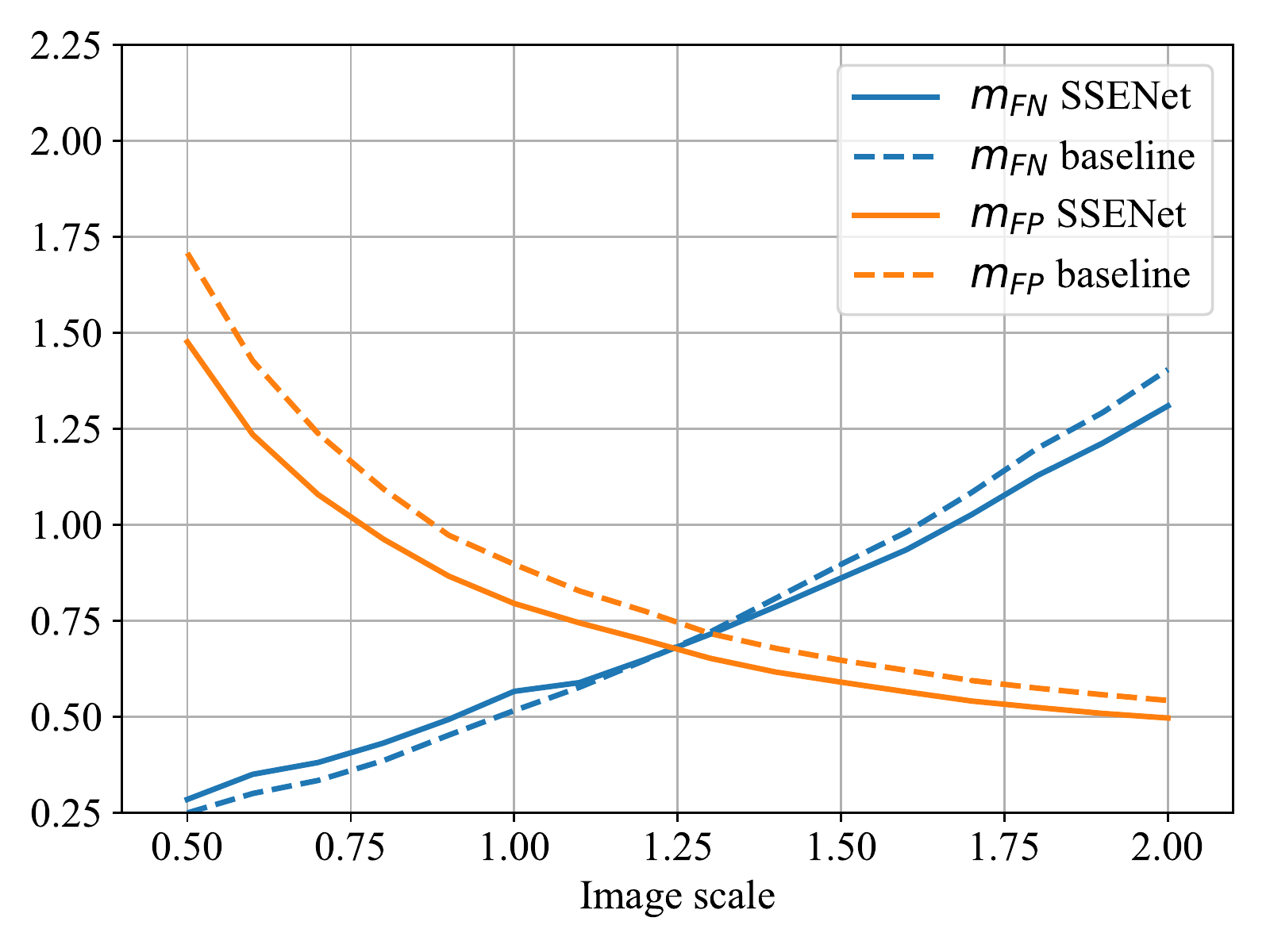}
		\caption{The model with scale equivariant regularization activates fewer background pixels (lower $m_{FP}$) and covers more object parts (lower $m_{FN}$ at right side) than traditional CAM method. The figure is best viewed on screen.}
		\label{fig:exp_a}
	\end{figure}
	\paragraph{Multi-scale Test} In many previous works, it is a common practice to compute multiple CAMs from multiple rescaled images and aggregate them to produce more accurate activation maps during test. In this paragraph, we evaluate SSENet and baseline model with both various single-scale test and multi-scale test. As shown in Fig.~\ref{fig:exp_b}, the CAMs of SSENet achieve higher mIoU than the baseline model with all kinds of single-scale test setting. Moreover, with multi-scale aggregation, our SSENet achieves a further improvement which also significantly beats the baseline. It demonstrates that our SSENet is effective for various scales and can be further improved with multi-scale aggregation.
	
	\begin{table*}[t]
		\centering
		\setlength{\tabcolsep}{0.7mm}{
			\begin{tabular}{lcccc}
				\hline
				Methods & Supervision & Saliency & mIoU(val) & mIoU(test)\\
				\hline
				EM-Adapt~\cite{papandreou2015weakly} & Image-level & - & 38.2\% & 39.6\%\\
				MIL~\cite{pinheiro2015image} & Image-level & - & 42.0\% & 40.6\%\\
				SEC~\cite{kolesnikov2016seed} & Image-level & - & 50.7\% & 51.7\%\\
				AffinityNet~\cite{ahn2018learning} & Image-level & - & \textbf{61.7\%} & \textbf{63.7\%}\\
				STC~\cite{wei2017stc} & Image-level & $\surd$ & 49.8\% & 51.2\%\\
				AdvErasing~\cite{wei2017object} & Image-level & $\surd$ & 55.0\% & 55.7\%\\
				SeeNet~\cite{hou2018self} & Image-level & $\surd$ & 63.1\% & 62.8\%\\
				DSRG~\cite{huang2018weakly} & Image-level & $\surd$ & 61.4\% & 63.2\%\\
				FickleNet~\cite{lee2019ficklenet} & Image-level & $\surd$ & 64.9\% & 65.3\%\\
				\hline
				What's Point~\cite{bearman2016s} & Point & - & 46.0\% & 43.6\%\\
				RAWK~\cite{vernaza2017learning} & Scribble & - & 61.4\% & -\\
				ScribbleSup~\cite{lin2016scribblesup} & Scribble & - & 63.1\% & -\\
				BoxSub~\cite{dai2015boxsup} & Bbox & - & 62.0\% & 64.6\%\\
				SDI~\cite{khoreva2017simple} & Bbox & - & 65.7\% & 67.5\%\\
				\hline
				\textbf{SSENet} & Image-level & - & \textbf{63.3\%} & \textbf{64.9\%}\\
				\hline
			\end{tabular}
		}
		\caption{Comparison with state-of-the-art weakly supervised approaches on both PASCAL VOC validation and test set. Our method is learned by image-level labels without extra supervisions.}
		\label{tab:voc}
	\end{table*}
	\paragraph{Source of Performance Improvement} Besides the visualization results shown in Fig.~\ref{fig:cams}, we prefer quantitative evaluations to investigate the source of performance improvement. Considering the evaluation metric is that
	\begin{equation}
	mIoU=\frac{1}{C}\sum_{c=0}^{C}\frac{TP_c}{TP_c+FN_c+FP_c},
	\end{equation}
	where $C$ is category number and $TP_c$, $FN_c$, $FP_c$ denotes the true positive, false negative, false positive predicted regions of each class respectively. The phenomenon in Fig.~\ref{fig:cams} shows that for small input images, the activation maps over-cover the object regions, leading to a lower proportion of $FN/FP$. When the input images are resized into large scale, the activation regions shrink into the most discriminative parts, causing a higher proportion of $FN/FP$. To further analyze the contributions of these two parts, we define another two metrics
	\begin{equation}\label{eq:mfn}
	m_{FN} =\frac{1}{C-1}\sum_{c=1}^{C-1}\frac{FN_c}{TP_c}.
	\end{equation}
	\begin{equation}\label{eq:mfp}
	m_{FP} =\frac{1}{C-1}\sum_{c=1}^{C-1}\frac{FP_c}{TP_c}.
	\end{equation}
	\begin{figure}[htbp]
		\centering
		\includegraphics[width=0.9\columnwidth]{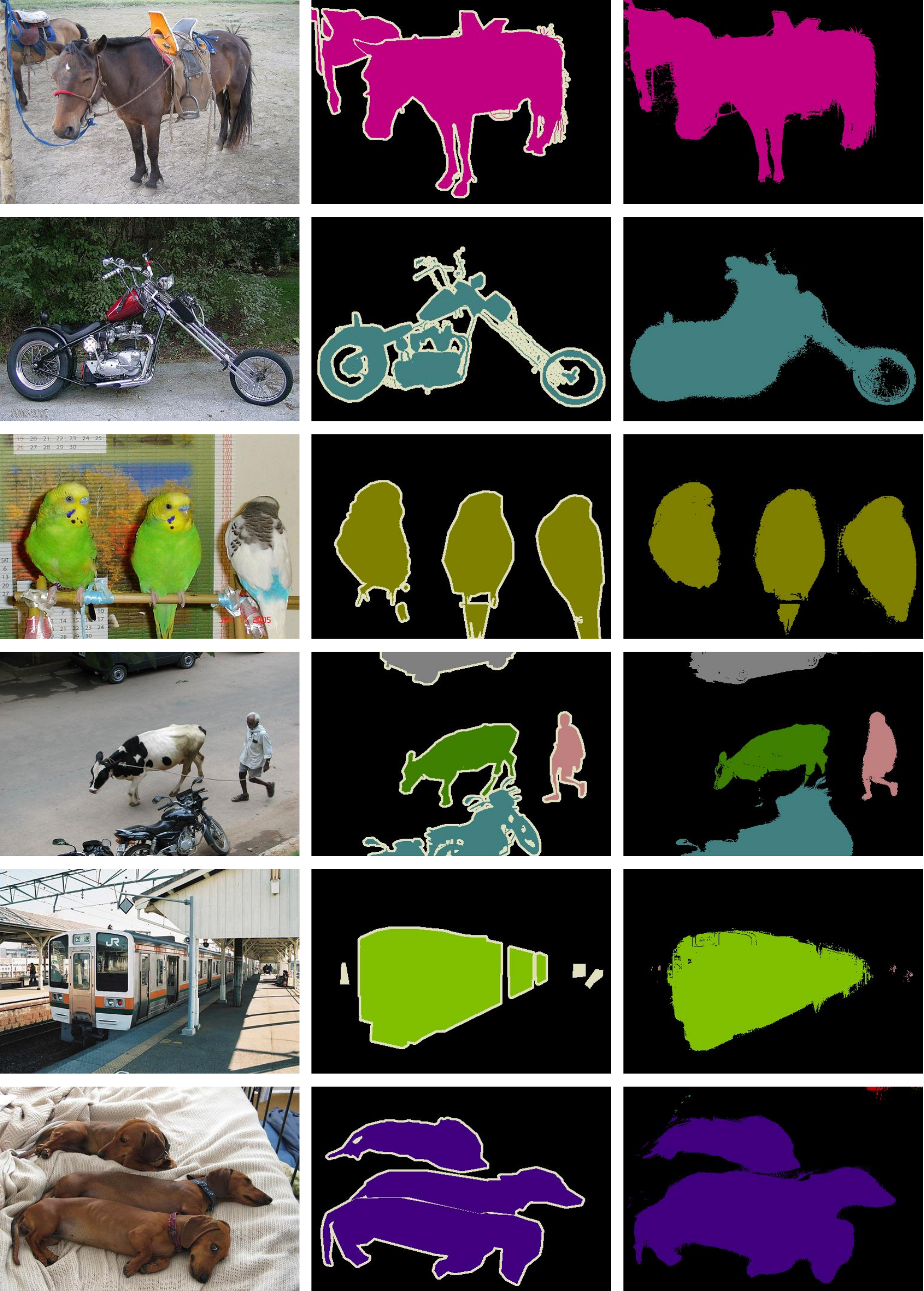}
		\caption{More segmentation results based on our approach. For each tuple, the left one is the original image, the middle is ground truth label and the right is the prediction of the final segmentation model based on our method. Our proposed weakly supervised method not only has complete segmentation coverage of large objects but also meets object boundary details.}
		\label{fig:seg}
	\end{figure}
	Note that the background category is excluded since the background activation region is reverse to the foreground categories. To simplify the analysis, we remove the background in these evaluation metrics. The Fig.~\ref{fig:exp_a} shows the curve of $m_{FN}$ and $m_{FP}$, based on the CAMs generated by the baseline model and SSENet in terms of various single-test scales. The tendency of these curves meets the claim that CNN generates rough CAM over-covering object when the input image is small, i.e., the prediction contains more false positive regions out of object regions. While the activation zone shrinks into the discriminative part with a larger input image and most parts of the object body are not activated, i.e., the prediction contains more false negative regions. Comparing the baseline and SSENet, the $m_{FP}$ curve demonstrates that the CAM generated by SSENet is more compact with fewer over-cover regions. Besides SSENet accurately activates more foreground regions only on large-size images since its $m_{FN}$ curve is higher than baseline at the right side of the axis while keeping lower at the left side. In shortly, the source of performance improvement mainly comes from less over-activated regions by scale equivariant regularization.	
	
	\subsection{Comparisons with State-of-the-arts}
	
	To further elevate the weakly supervised semantic segmentation network performance, we follow the work of Affinity~\cite{ahn2018learning} to expand and constrain the class activation maps achieved by our proposed SSENet. Moreover, we reimplement DeepLab with ResNet38 as the backbone, using modified activation maps as pseudo labels for semantic segmentation training. The Tab.~\ref{tab:voc} illustrates that the final result of our SSENet has significant improvement than AffinityNet baseline. The performance elevation mainly stems from the improved CAMs and the more accurate pseudo segmentation labels. Besides, the method based on SSENet even beats some advanced approaches which are embedded with additional off-the-shelf saliency methods. Moreover, our method also achieves comparable performance with the state-of-the-art weakly supervised semantic segmentation approaches based on stronger supervisions like point, scribble and bounding box.
	
	\section{Conclusion}
	In this paper, we resort to the self-supervision regularization for weakly supervised semantic segmentation. We propose scale equivariant regularization (SER) to deal with the inconsistency of network activation map on various image sizes. With the SER, the class activation maps from scale augmented images keep the same after rescaled into the same size. Based on this regularization, we design a two-branch self-supervised scale equivariant network (SSENet) for class activation map learning with only image-level supervision. The network learns more discriminative regions on large objects and overcomes the phenomenon of over-activated on small objects. We evaluate the proposed method on PASCAL VOC 2012 dataset and the results demonstrate that our approach has outstanding performance than other state-of-the-art weakly supervised methods.
	
	\bibliographystyle{aaai}
	\bibliography{ser}

\end{document}